\documentclass[journal]{IEEEtran}
\usepackage{amsmath}
\usepackage{graphicx}
\usepackage{multirow}
\usepackage{subfigure}
\usepackage{makecell}
\usepackage{color}
\usepackage{booktabs}
\usepackage{epstopdf}
\usepackage{algorithm}
\usepackage{algorithmicx}
\usepackage{algpseudocode}
\usepackage{url}
\usepackage{amsmath}
\usepackage{amssymb}  
\usepackage{pifont}

\usepackage[colorlinks,
]{hyperref}
\usepackage{graphicx}
\usepackage{subfigure}
\usepackage{amsmath}
\hypersetup{hidelinks}
\usepackage{cite} 
\usepackage{threeparttable} 
\usepackage{hyperref}
\usepackage{graphicx}

\usepackage{orcidlink}

\usepackage{xcolor}
\usepackage{hyperref} 

\definecolor{myblue}{RGB}{0, 0, 255}   

\hypersetup{
  colorlinks=true,
  linkcolor=myblue,  
  citecolor=myblue, 
  urlcolor=myblue
}

\begin{document}
%
\title{EraW-Net: Enhance-Refine-Align W-Net for Scene-Associated Driver Attention Estimation}

\author{%
    \IEEEauthorblockN{
        \mbox{Jun Zhou}$^{\orcidlink{0009-0002-4135-7869}}$,~%
        \mbox{Chunsheng Liu$^{\orcidlink{0000-0001-5516-2486}}$,~\IEEEmembership{Member, IEEE}},~%
        \mbox{Faliang Chang}$^{\orcidlink{0000-0003-1276-2267}}$,~%
        \mbox{Wenqian Wang}$^{\orcidlink{0000-0003-0285-9786}}$,~%
    }
    
    \IEEEauthorblockN{
        \mbox{Penghui Hao}$^{\orcidlink{0009-0007-0398-506X}}$,~%
        \mbox{Yiming Huang}$^{\orcidlink{0009-0005-8246-7652}}$,~
        \mbox{Zhiqiang Yang}$^{\orcidlink{0009-0005-0387-2028}}$ 
    }

\thanks{This work was supported in part by the National Natural Science Foundation of China under Grant U22A2058, Grant 62176138, and Grant 62176136, in part by National Key R \& D Program of China under Grant 2018YFB1305300, in part by the Shandong Outstanding Youth Funding under Grant ZR2023YQ054, and in part by the Shandong Provincial Key Research and Development Program (Major Scientific and Technological Innovation Project) under Grant 2019JZZY010130 and Grant 2020CXGC010207. (Corresponding author: Chunsheng Liu; Faliang Chang.)}   
\thanks{Jun Zhou, Chunsheng Liu, Faliang Chang, Penghui Hao, Yiming Huang and Zhiqiang Yang are with the School of Control Science and Engineering, Shandong University, Jinan 250061, China(email: zhoujuner@mail.sdu.edu.cn; liuchunsheng@sdu.edu.cn; flchang@sdu.edu.cn; haopenghui@mail.sdu.edu.cn; 202320747@mail.sdu.edu.cn; yangzhiqiang@mail.sdu.edu.cn).}
\thanks{Wenqian Wang is with the School of Electrical and Electronic Engineering, Nanyang Technological University, Singapore 639798, Singapore (e-mail: wenqian.wang@ntu.edu.sg).}
}

\markboth{}
{Shell \MakeLowercase{\textit{et al.}}: Bare Demo of IEEEtran.cls for IEEE Journals}

%



\maketitle


\begin{abstract}
Associating driver attention with driving scene across two fields of views (FOVs) is a challenging cross-domain perception problem, which requires comprehensive consideration of cross-view mapping, dynamic driving scene analysis, driver status tracking. Previous methods typically focus on a single view or map attention to the scene via estimated gaze, failing to exploit the implicit connection between them. Moreover, simple fusion modules are inadequate for modeling the complex relationships between the two views, complicating information integration. To address these issues, we propose EraW-Net, a novel end-to-end method for scene-associated driver attention estimation. This method enhances the most discriminative dynamic cues, refines feature representations, and facilitates semantically aligned cross-domain integration through a W-shaped architecture, termed W-Net. Specifically, a Dynamic Adaptive Filter Module (DAF-Module) is proposed to address the challenges of frequently changing driving environments by extracting vital regions. It suppresses the indiscriminately recorded dynamics and highlights crucial ones by innovative joint frequency-spatial analysis, enhancing the model's ability to parse complex dynamics. Additionally, to track driver states during non-fixed facial poses, we propose a Global Context Sharing Module (GCS-Module) to construct refined feature representations by capturing hierarchical features that adapt to various scales of head and eye movements. Finally, W-Net achieves systematic cross-view information integration through its ``Encoding-Independent Partial Decoding-Fusion Decoding” structure, addressing semantic misalignment in heterogeneous data integration. Experiments demonstrate that the proposed method robustly and accurately estimates driver attention mapping in scenes on large public datasets.

\end{abstract}


%
\IEEEpeerreviewmaketitle

\begin{figure*}[t]
\centering
\includegraphics[width=16cm]{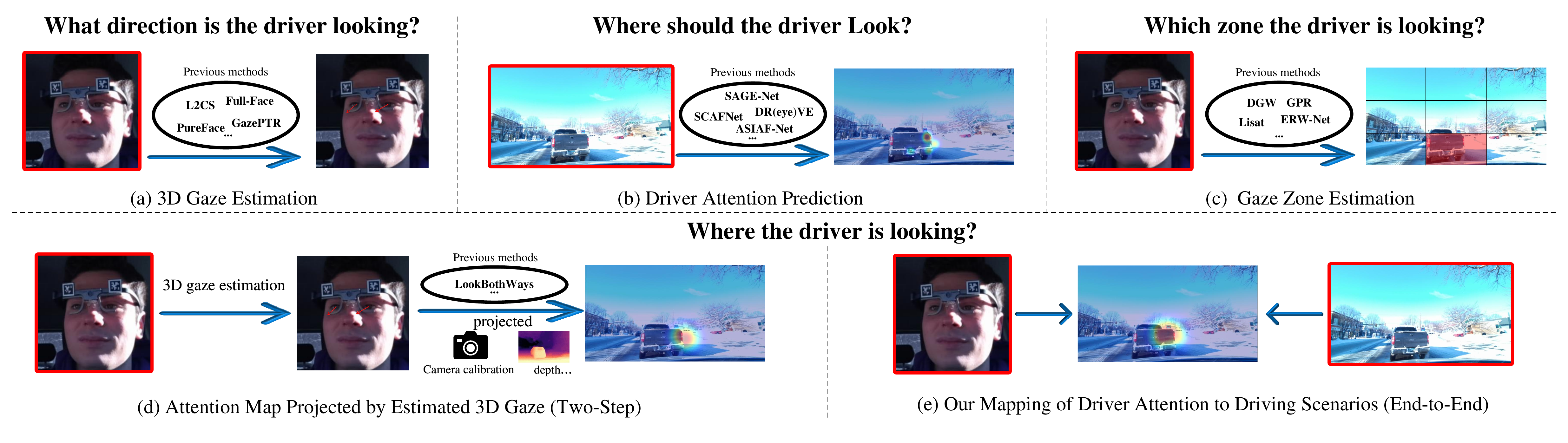}
\caption{\textbf{Comparison of different tasks related to driver attention.} The inputs to the model are highlighted with red boxes in the figure. \textbf{(a) 3D gaze estimation:} Determine what direction the driver is looking. \textbf{(b) Driver attention prediction:} Predict where should the driver look. \textbf{(c) Gaze zone estimation:} Estimate which zone the driver is looking. \textbf{(d)} and \textbf{(e)} are both for estimating where the driver is looking in the current scene. \textbf{(d) Attention map projected by estimated 3D gaze:} The results of 3D gaze estimation are mapped onto the scene image using camera parameters and depth information, which is a two-step estimation method. \textbf{(e) Our proposed end-to-end mapping of driver attention to driving scene:} End-to-end estimate driver attention in the current environment by leveraging complementary information from both driver-facing and scene-facing views. 
}
\label{fig:idea}
\end{figure*}

\section{Introduction}
\IEEEPARstart{D}river monitoring systems \cite{mou2021isotropic}, \cite{wang2023100} 
have gained widespread attention in recent years due to their ability to perceive the driver's state in real-time and adjust behavior accordingly. Among the various research topics within this domain, the association between driver attention and the driving scene stands out as particularly valuable, focusing on estimating regions or targets within the current environment that are being attended to by the driver. Such studies have vast potential applications in enhancing road safety \cite{xu2021improvement}, improving human-vehicle interaction \cite{capallera2022human}, and supporting assisted driving decision-making processes \cite{han2024dme}.

This research addresses the complex problem of understanding the driving scene, capturing driver attention, and mapping this attention to the scene. The primary challenges in this domain stem from the fact that driver attention and the road condition correspond to two distinct perspectives: the interior (in-vehicle) and the exterior (on-road) views. There is no explicit overlapping information between them, making the cross-domain integration of information a highly demanding task, and the inherent complex causal relationships between these perspectives further exacerbate this challenge. Specifically, driver attention is heavily influenced by traffic conditions. However, due to the driver's subjective initiative, their attention may shift multiple times even before any noticeable changes occur in the driving environment. This makes it extremely difficult to clearly define the relationship between shifts in driver attention and changes in the driving environment. Furthermore, there are continuous and unpredictable changes in both the driver's behavior and the road condition during driving tasks. Concretely, drivers frequently engage in substantial eye movements and head rotations to gather information, complicating the precise tracking of their attention due to these dynamic postures. Additionally, sudden occurrences in the driving environment require immediate responses, such as the appearance of pedestrians or lane changes by other vehicles, challenging the adaptability of monitoring models. 

Many studies have focused on analyzing a single perspective. For instance, attention estimation from the driver's perspective captures rapidly changing states \cite{yang2020driver} of the driver by analyzing a composite \cite{cheng2024you} of facial expressions \cite{zhang2024domain}, head movements \cite{lu2023jhpfa} and other features \cite{yang2021driver}, 
as shown in \textbf{Fig. \ref{fig:idea}(a)}. On the other hand, some studies predict areas 
\cite{fang2021dada},  \cite{lv2020improving} or targets 
\cite{li2022adaptive}, \cite{qin2022id} that drivers should attend to by analyzing key traffic elements \cite{pal2020looking} from the road perspective, as shown in \textbf{Fig. \ref{fig:idea}(b)}. However, these studies are limited to single-perspective analysis and do not achieve the integration of both.

In recent years, several studies have attempted to model the association between driver attention and the scene at the result level. Some work, as shown in \textbf{Fig. \ref{fig:idea}(c)} has focused on mapping driver attention to different cockpit regions \cite{vora2018driver}, \cite{shah2022driver} or directly estimating the driver’s point \cite{dua2020dgaze} of gaze on the road by detecting the driver’s state. These methods overlook the complementary information \cite{du2023causes} between the two perspectives. Kasahara et al. \cite{kasahara2022look} proposed a self-supervised 3D geometric learning framework to enhance the geometric consistency between driver gaze and scene saliency. Hu et al. \cite{hu2021data} developed a driver attention estimation model based on scene image information and probability maps generated by gaze vectors, as shown in \textbf{Fig. \ref{fig:idea}(d)}. However, these methods also fail to fully utilize the complementary information between the two perspectives.

Modeling complex relationships and exploiting implicit connections between the two perspectives are crucial for associating driver attention with the scene. To leverage information from two data sources, existing methods primarily rely on feature fusion modules. However, applying these methods to our task presents several challenges: (1) The cross-view information from the driver and the driving scene is highly complex, making simple fusion insufficient to capture all essential features and relationships; (2) Fusion methods primarily focus on the relationships between features but do not guarantee the quality of the individual features involved in the fusion process; (3) Fusion methods cannot resolve semantic inconsistencies during fusion, limiting the model's performance. To address these issues, we have developed a W-shaped network structure, named W-Net. W-Net employs an ``Encoding-Independent Partial Decoding-Fusion Decoding" architecture to systematically aggregate complementary information from both inputs. It effectively learns functional features from two inputs, models their complex relationships, and enhances the model's stability and performance, rather than merely merging information.



In this paper, to bridge the gap of correlating driver attention with the scene in real dynamic driving tasks, we propose the Enhance-Refine-Align W-shaped Network (EraW-Net). EraW-Net enhances discriminative visual cues, refines feature representations, and integrates cross-domain information with semantic alignment, addressing issues of associating data from separate perspectives and adapting to complex dynamic environments and frequent driver posture changes.

Specifically, 
to overcome the limitations of basic fusion methods in capturing essential features and resolving prevalent semantic inconsistencies, we explore the W-Net architecture to systematically model the complex relationships in cross-domain information. It adopts a W-shaped design and follows an ``Encoding-Independent Partial Decoding-Fusion Decoding" structure. W-Net supports two parallel inputs, making it widely applicable to cross-domain integration tasks. This structure focuses on an innovative Two-Stage Decoding (TS-Decoding) strategy, which independently decodes each feature before merging them, reinforcing features within each domain and ensuring intra-domain semantic alignment especially. Besides, we employ cross-domain attention mechanisms to fuse features from both domains and utilize our designed Channel-Space Hybrid Attention (CSHA) units for layer-by-layer decoding. Our proposed method introduces a new architecture for cross-domain feature integration in image generation tasks, effectively overcoming the challenge of semantic misalignment in the integration of heterogeneous data sources.

To extract the vital dynamics in frequently changing driving environments, we propose the Dynamic Adaptive Filter Module (DAF-Module) for feature enhancement. Unlike traditional methods that directly use optical flow as one branch \cite{palazzi2018predicting} to handle dynamic information, we calculate inter-frame dynamic features by computing the correlation of encoded features from the same layer of adjacent frames. This approach improves information utilization efficiency and reduces computational costs. The module focuses on innovative joint frequency-spatial analysis, where the former learns informative representations of global receptive fields, filtering out indiscriminately redundant recorded frequency components, while the latter enhances spatial areas with significant short-term changes, complementing each other. This processing enhances the model's robustness and adaptability in complex dynamic environments.

To tackle the challenge of tracking driver states during non-fixed facial poses, we propose the Global Context Sharing Module (GCS-Module) for refining facial feature representations. Unlike previous approaches that construct shared feature sets \cite{martin2018dynamics} as inputs, our method focuses on extracting and aggregating multi-scale features to refine feature representations. This representation encompasses information from local details to global structures, enabling adaptation to varying scales of features such as significant head poses and subtle eye movements, thereby enhancing the accuracy of state recognition.

Our contributions can be summarized as follows:

1) We are the first to achieve end-to-end mapping of driver attention to the driving scene, pioneering the exploitation of the implicit connection and modeling the relationship between them. We explore the W-Net architecture for addressing the semantic misalignment challenge in integrating heterogeneous data from two sources.
 
2) To extract vital motion driven by continuous and varied stimuli, we propose the DAF-Module for feature enhancement through joint frequency-spatial filtering, enhancing the model’s ability to adapt to continuous dynamics. Additionally, the GCS-Module is designed to address the challenge of tracking driver states with variable facial poses. It leverages hierarchical features to establish a refined global multi-scale representation of the driver’s face.

3) Experiments conducted on a large-scale public dataset demonstrate that the proposed method accurately estimates the driver’s pixel-level attention mapping in driving scenes, outperforming existing methods.

The paper is organized as follows. Section II provides a brief introduction of the related methods. Section III describes the proposed EraW-Net methods. Section IV presents evaluation and comparison results, and Section V concludes with future work.

\section{Related Work}
\subsection{Driver Attention Estimation}

The essence of driver attention estimation centers on remote gaze tracking, aiming to determine gaze zones, 3D gaze direction, or fixation points. However, traditional methods relying on near-infrared (IR) illumination technology exhibit limited robustness due to varying lighting conditions in driving environments \cite{naqvi2018deep}. With the rapid advancement of deep learning techniques, vision-based methods have gained favor in the academic community due to their affordability, ease of data acquisition, and high accuracy.

Remote gaze tracking in outdoor environments presents significant challenges. Researchers often address this issue by estimating the driver's gaze area within the cockpit. Most related datasets divide the cockpit environment into 6-9 Areas of Interest (AOIs) 
\cite{ghosh2021speak2label}, \cite{vora2018driver} to achieve a coarse estimation of attention. However, the complexity of the driving environment and substantial variations in the driver's head pose make it difficult to distinguish between adjacent areas. Numerous studies have improved estimation accuracy by constructing composite feature sets that combine facial landmarks \cite{yuan2022self}, eye features \cite{yang2021driver}, and head pose \cite{wang2019continuous}. While these methods offer rough estimations of the driver's attention, some applications require higher precision. This approach encounters significant challenges when applied to finer-grained divisions.

To address this issue, some research resolves this issue by intersecting the driver's gaze direction with the plane of the road, which rely on accurate 3D gaze. Most methods use cropped facial images as input for 3D gaze estimation. Various models 
\cite{balim2023efe}, \cite{chen2022towards} have been proposed to improve gaze estimation accuracy, effectively handling challenges such as lighting variations and facial occlusions. These models have demonstrated their effectiveness in synthetic data \cite{jindal2023cuda}, laboratory settings 
\cite{zhang2020eth} and in-the-wild environments \cite{kellnhofer2019gaze360}.

However, in driving environments, even minor errors in 3D gaze estimation can be significantly amplified due to long projection distances. To address this issue, Bhagat et al. \cite{bhagat2023driver} proposed a Point of Gaze (PoG) estimation system based on existing gaze estimation methods to better understand the interaction between the driver and the driving environment. Jha et al. \cite{jha2022estimation} developed a probabilistic model-based approach to create salient regions describing a driver's visual attention. Additionally, Dua et al. \cite{dua2020dgaze} created the first driver gaze mapping dataset recorded using a smartphone camera and introduced the I-DGAZE model to predict driver gaze on the road end-to-end. While these methods establish a connection between driver attention and the driving scene, they still face challenges in terms of accuracy and interpretability.

\subsection{Joint Monitoring Based on Driver Status and Road Scene}

There is a close correlation between a driver behavior and the surrounding driving environment. In response, numerous innovative studies have emerged that integrate driver's behavior with traffic environment information for enhanced monitoring. This integration significantly improves the accuracy of predicting driving intentions, operational actions, and attention estimation in driver monitoring systems.

For example, 
Rong et al. \cite{rong2020driver} analyzed motion features in traffic scenes and combined them with dynamic in-vehicle information, using video data from both inside and outside the car to detect driver intentions. Huang et al. \cite{huang2024driver} proposed a topological graph method based on the interaction between driver behavior and the traffic environment to predict driving intentions, highlighting the importance of behavior-environment interaction. Yang et al. \cite{yang2023real} developed a decision-level fusion architecture that utilized EEG signals, eye movements, and vehicle status data to infer the driver’s workload. Konrad et al. \cite{konrad2023classification} integrated driver head direction data with vehicle perception systems to determine whether the driver was observing surrounding objects, thereby effectively adjusting warning systems or autonomous driving interventions. \cite{kasahara2022look} employed a self-supervised method to enhance the geometric consistency between the driver’s gaze and the scene saliency, resulting in more accurate estimations.

In summary, these research findings highlight the effectiveness of integrating driver state information with the driving scene for enhanced driver monitoring. However, in the field of driver attention estimation, there has been a notable gap in research that integrates data from both the driver and the environment at the feature level. Our work aims to address this gap by exploring the integration of these two data sources more comprehensively.

\begin{figure}[t]
\centering
\includegraphics[width=9cm]{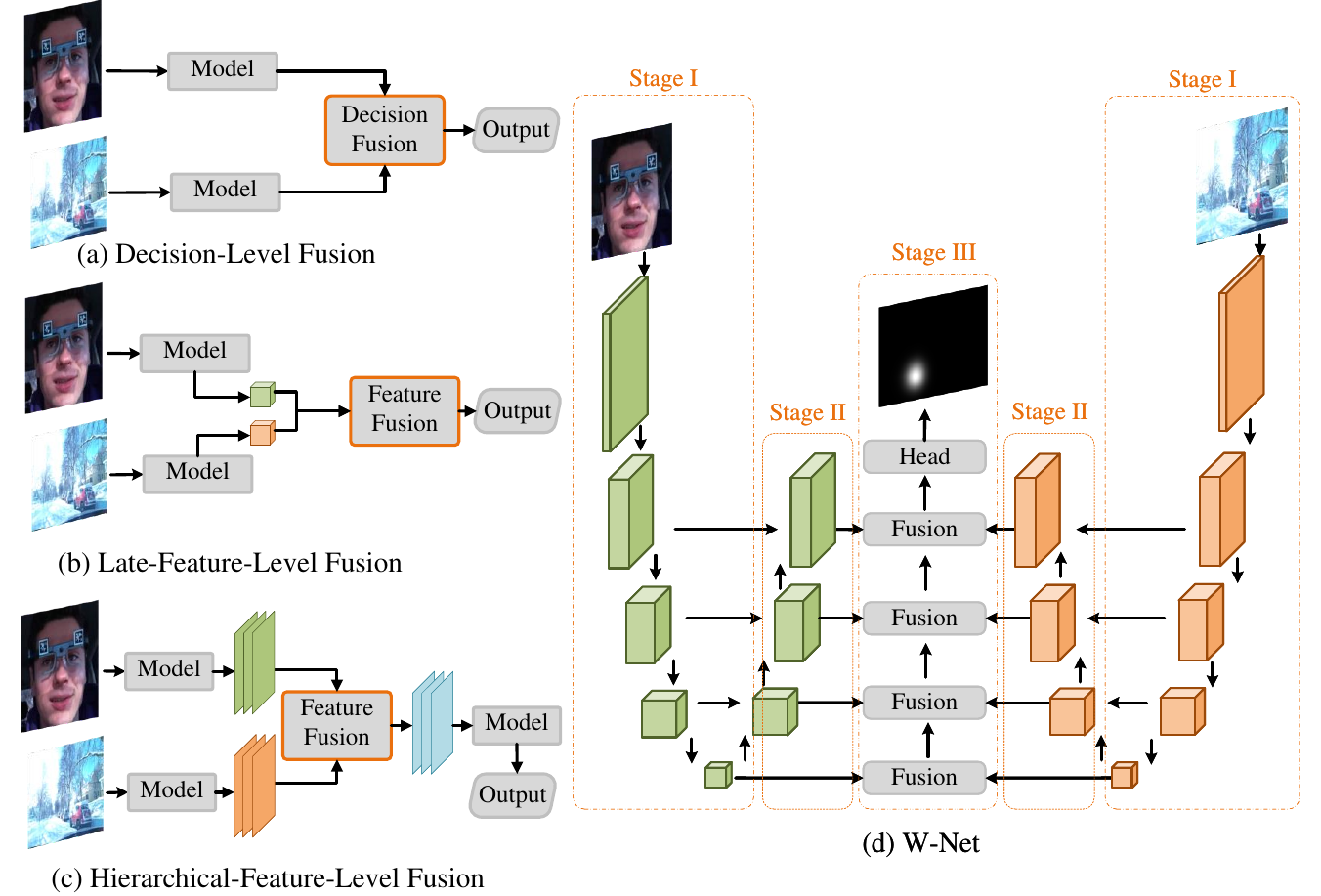}
\caption{Comparison of three fusion strategies and our W-Net for cross-domain information integration. (a) Decision-Level Fusion \cite{wang2023crack}: Aggregate the independent decision outputs from the two sources into a final decision. (b) Late-Feature-Level Fusion 
\cite{liu2023cross}: Process each input independently, then integrate the late features through post-processing. (c) Hierarchical-Feature-Level Fusion 
\cite{wen2023msgfusion}: Fuse features from both inputs layer-by-layer. (d) W-Net: Use an architecture of ``Encoding(Stage I)-Independent Partial Decoding(Stage II)-Fusion Decoding(Stage III)" to integrate information, support for two inputs from two different domains. Details are provided in Section III.D. }
\label{fig:fusion classification}
\end{figure}

\begin{figure*}[t] 
\centering
\includegraphics[width=16cm]{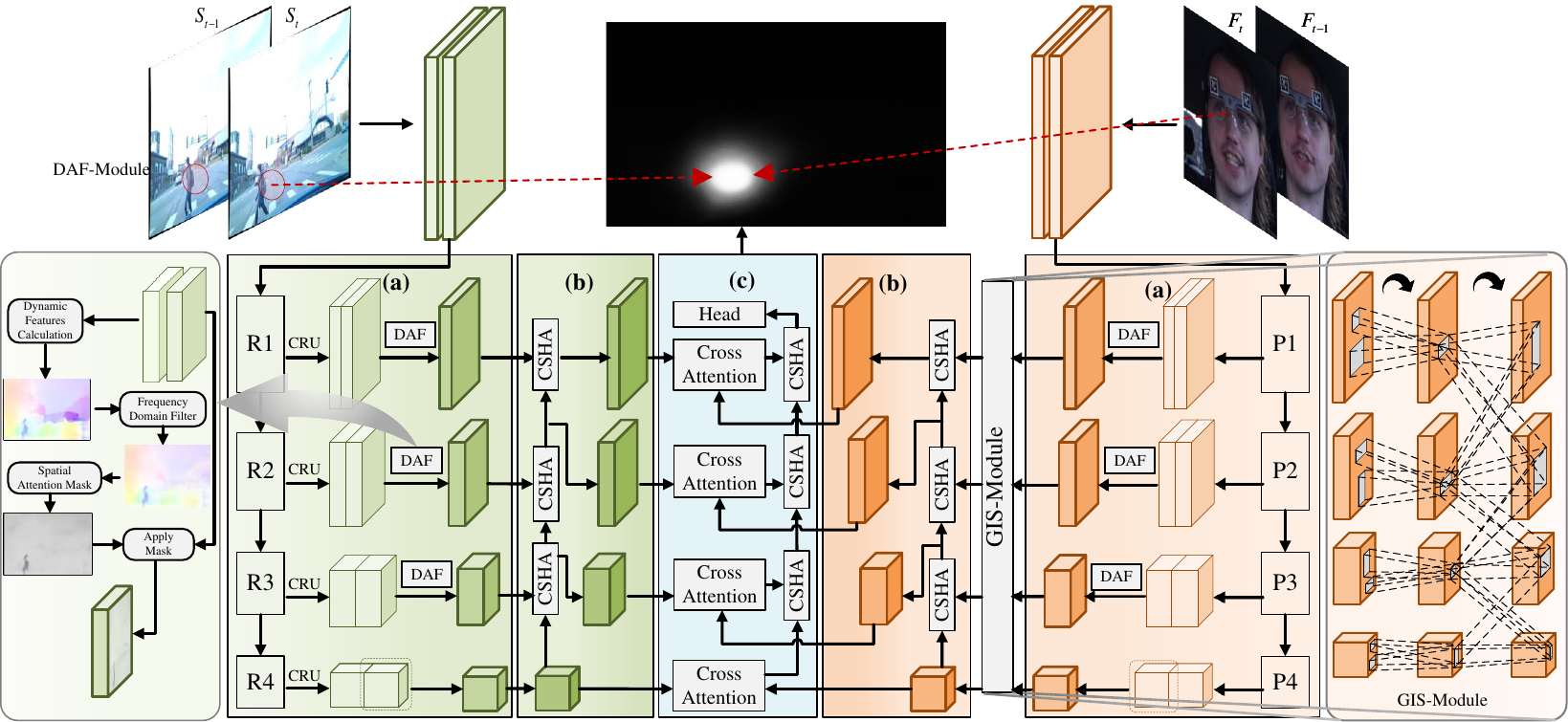}
\caption{Architecture of EraW-Net. The overall architecture is based on the proposed W-Net structure, which includes three key stages: (a) encoding, (b) independent partial decoding, and (c) fusion decoding. During feature encoding, Channel Reduction Units (CRUs) are employed to standardize channel dimensions across corresponding layers of both branches. The Dynamic Adaptive Filter Module (DAF-Module) employs joint frequency-spatial filtering masks derived from inter-frame dynamics to emphasize significant dynamics within original features. The Global Context Sharing Module (GCS-Module) extracts and consolidates multi-scale features to refine a comprehensive global representation of facial features. The core of W-Net's Two-Stage Decoding (TS-Decoding) strategy (depicted in (b) and (c)) lies in its approach of intra-domain feature alignment before fusion decoding, ensuring semantic consistency across domains. This methodology ensures that subsequent fusion processes operate on well-aligned and high-quality feature representations.
}
\label{fig:architecture}
\end{figure*}

\begin{figure}[t]
\centering
\includegraphics[width=8cm]{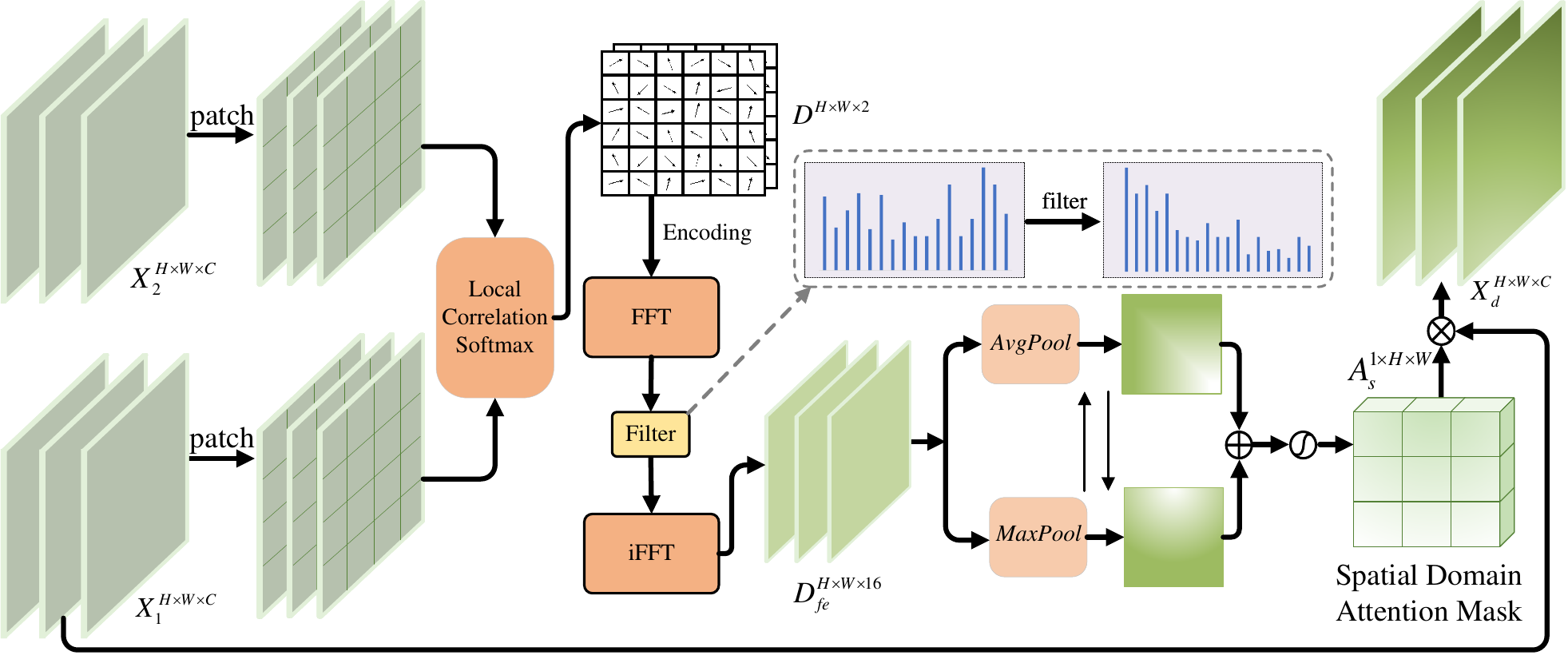}
\caption{The Dynamic Adaptive Filter Module (DAF-Module) processes inter-frame dynamic information by innovative joint
frequency-spatial analysis, guiding the model to focus on critical motion characteristics. It first calculates dynamic features through local correlation, filters out redundant dynamics in the frequency domain. and then enhances spatial areas with significant short-term changes to enhance the features representation.}
\label{fig:3DV2}
\end{figure}

\subsection{Cross-modal representation learning}
Given the potential presence of complementary information from two modalities within a single task, cross-modal representation learning aims to understand and represent these dual data sources through machine learning. This domain primarily relies on feature fusion strategies to combine cross-modal information. Some straightforward fusion methods 
\cite{liu2023cross}, \cite{li2020spatial} employ ``Decision-Level Fusion" (\textbf{Fig. \ref{fig:fusion classification} (a)}) and ``Late Feature-Level Fusion" (\textbf{Fig. \ref{fig:fusion classification} (b)}) strategies, using element-wise multiplication, addition, or concatenation operations to obtain fused features. To leverage the advantages of both early and late fusion, various models 
\cite{fan2024querytrack}, \cite{xiao2023dgfnet} have proposed ``Hierarchical Fusion" (\textbf{Fig. \ref{fig:fusion classification} (c)}) of features from two data sources, enabling shared features and information exchange. This approach deepens the understanding of the relationship between the two data sources. However, the fusion-based methods mentioned have notable limitations. Specifically, they focus solely on the fusion process without guarantying the quality of the features being fused. Moreover, semantic misalignment issues occur during fusion, and these problems can originate from within the modalities themselves prior to the fusion process. To address these issues, we explore a W-Net architecture (\textbf{Fig. \ref{fig:fusion classification} (d)}) to systematically integrate information from two sources. In particular, it performs intra-modal semantic alignment before feature fusion, thereby reducing the potential for information confusion that can result from direct modality fusion. This architecture offers an innovative solution to the challenges associated with cross-modal data processing.

\section{Methods}

In this section, we introduce the Enhance-Refine-Align W-shaped Network (EraW-Net), which is proposed to associate driver attention with driving scenes by integrating information from distinct perspectives. The EraW-Net model comprises three key components: (1) the DAF-Module is proposed to address the challenges posed by continuous and uncertain movements during driving. This module enhances feature representation by leveraging inter-frame motion information and directs the model to focus on discriminative cues through joint filtering in both frequency and spatial domains; (2) the GCS-Module is introduced to tackle the issue of capturing attention despite the driver's non-fixed poses. It aggregates multi-scale features both within and between layers to adapt to different levels of critical information such as head rotations and eye movements, thereby refining facial feature representation; and (3) a W-Net architecture is explored to overcome the issues of insufficient feature extraction and semantic inconsistency in cross-domain information integration. The main modules are detailed in the following subsections.

\subsection{Overall architecture}
The overall architecture of EraW-Net is based on our proposed ``Encoding-Independent Partial Decoding-Fusion Decoding" framework, as illustrated in \textbf{Fig. \ref{fig:architecture}}. Initially, two branches independently encode the driver's facial image and the scene image to extract contextual spatial information. The DAF-Module is applied to multi-level features of both the face and scene to enhance dynamic regions. To accurately capture the driver's attention state, the GCS-Module is applied exclusively to the driver's facial image, decoding the features step-by-step after fusing them with the dynamically processed attention mask. Subsequently, a TS-Decoding strategy is employed: first, independent partial decoding is performed on the features from each branch; then, the decoded features are fed into the cross-domain feature integration stage, where multi-level features are progressively decoded. This process ultimately generates a high-resolution driver attention heatmap.

\subsection{DAF-Module Based Feature Enhancement}
The facial feature extraction branch processes two consecutive facial images $F_{t}^{3 \times H_F \times W_F}$ and $F_{t-1}^{3 \times H_F \times W_F}$ as input, where $H_F = W_F = 224$. The ResNet-18 architecture, excluding its final fully connected layer, serves as the backbone network for feature extraction. The backbone is divided into four stacked convolutional units $(P_1, P_2, P_3, P_4)$, with the output features from each unit denoted as $(f_{i, t}, f_{i, t-1})$, $i \in \{1, 2, 3, 4\}$, where $f_{i, t}$ and $f_{i, t-1}$ have have dimensions $C_{Fi} \times H_{Fi} \times W_{Fi}$,$C_{Fi} = 64 \times 2^{i-1}$, $H_{Fi} = \frac{H_F}{2^{i+1}}$, and $W_{Fi} = \frac{W_F}{2^{i+1}}$.

The traffic scene feature extraction branch utilizes synchronized traffic scene image frames $S_t^{3 \times H_S \times W_S}$ and $S_{t-1}^{3 \times H_S \times W_S}$ as inputs, where $H_S = 480$ and $W_S = 800$. The ResNet-50 architecture serves as the backbone network for feature extraction and is segmented into four stacked convolutional units $(R_1, R_2, R_3, R_4)$. The output features from each unit are represented as $(s_{i, t}, s_{i, t-1}$, $i \in \{1, 2, 3, 4\}$, where $s_{i, t}$ and $s_{i, t-1}$ have dimensions ${C_{Si} \times H_{Si} \times W_{Si}}$,$C_{Si} = 64 \times 2^{i+1}$, $H_{Si} = \frac{H_S}{2^{i+1}}$, and $W_{Si} = \frac{W_S}{2^{i+1}}$. The channel dimensions of these feature maps are then standardized to $C_{Si} = C_{Fi} = 64 \times 2^{i-1}$ through corresponding channel reduction units (CRU), 
\begin{equation}
\begin{aligned}
CRU(a_i) &= Conv_{3 \times 3} \left( Cat\left[ Conv_{1 \times 1} (a_{i}), \right. \right. \\
&\phantom{{}= Conv_{3 \times 3} \left( Cat\left[\right.\right.} \left. \left. Conv_{3 \times 3} (Conv_{1 \times 1} (a_{i})) \right] \right)
\end{aligned}
\end{equation}
where, $a _ { i }$ represents the input features and  $a _ { o }$ represents the output features. The operator $Cat ( \cdot )$  denotes feature concatenation. $Conv_ { 1 \times 1 } ( \cdot )$ and $Conv_ { 3 \times 3 } ( \cdot )$ represent convolution operations with  $1 \times 1$ and $3 \times 3$ kernels, respectively.

\textbf{DAF-Module}: Unlike previous approaches that incorporate optical flow information as an independent branch input to the network, our method calculates inter-frame dynamics at the same hierarchical level from the encoded features of adjacent frames. Through frequency domain filtering and spatial attention, redundant information is removed, allowing the model to focus on the most discriminative features. The structure of the module is shown in \textbf{Fig. \ref{fig:3DV2}}.

Firstly, we divide the input features of adjacent frames $(X_{1}, X_{2})$ with dimension of ${H \times W \times C}$ into non-overlapping feature subregions $(L_{n1}, L_{n2})$ with dimention of ${H_L \times W_L \times C}$, where $n \in \{1, 2, \dots, \frac{H}{H_L}\times \frac{W}{W_L}\}$, where, $H_L$ and $W_L$ denote the height and width of the subregions, respectively. Within each subregion, we independently perform correlation calculations, followed by a weighted average with the pixel grid $P_{Ln} \in \mathbb R^{H \times W \times 2}$. This step determines the optimal local motion vector for each pixel, denoted as $\tilde {P}_{Ln}$. The process is described by the following equation:
\begin{equation}
    Corr_{Ln} = \frac{{L_{n1}} \cdot {L_{n2}^T}}{\sqrt{C}}
\end{equation}
\begin{equation}
    \tilde {P}_{Ln} =  softmax(Corr_{Ln}) \times {P}_{Ln}
\end{equation}

Each element in $Corr_{Ln} \in \mathbb {R}^{H_L \times W_L \times H_L \times W_L}$ represents the correlation value between coordinates in $L_{n1}$ and $L_{n2}$.
Finally, by calculating the difference from ${P}_{Ln}$, we obtain the dynamic feature map $D_{Ln} \in \mathbb R^{H_L \times W_L \times 2}$ of the subregion.
Following this, by integrating the dynamic features predicted for all local blocks, we obtain an overall dynamic feature estimate, denoted as $D \in \mathbb {R}^{H \times W \times 2}$. It is encoded to derive a higher-level semantic representation $D_e \in \mathbb {R}^{H \times W \times 16}$.
In the above process, motion information from all regions is recorded. To emphasize key dynamic information in the image, we apply frequency domain filtering. 
This procedure can be summarized as:
\begin{equation}
    D_{fe} = IFFT(Conv_{1\times1}((FFT(D))
\end{equation}
where, $FFT$ is Fast Fourier Transform and $IFFT$ is Inverse Fast Fourier transform.

We input $D_{fe}$ into a spatial attention unit, denoted as $SA(\cdot)$. The process is described as:

\begin{equation}
    A_{s} =\sigma(Conv_{7 \times 7}(Cat\left[\frac{1}{C}\sum_{c=1}^{C}D_{fe}, \underset{c=1}{\overset{C}{\max}}(D_{fe})\right]))
\end{equation}
where,
$\sigma$ is the sigmoid function, and $A_s \in \mathbb {R}^{1 \times H \times W}$.

Then, we apply this attention mask to $X_{1}$ to further extract dynamic spatial features $X_{d} \in \mathbb R^{H\times W\times C}$.
We apply the DAF-Module separately to the driver's facial features and scene features:
\begin{equation}
f _ { d i }  = \begin{cases} 
f _ { i t }+ DAF(f _ { i t } , f _ { i - 1 })&, i \in \{1,2,3\}, \\
f _ { i t }  &,i = 4 
\end{cases}
\end{equation}
\begin{equation}
s _ { d i }  = \begin{cases} 
s _ { i t }+ DAF(s _ { i t } , s _ { i - 1 })&, i \in \{1,2,3\}, \\
s _ { i t }  &,i = 4 
\end{cases}
\end{equation}
where, $f_{di}$, $f_{i, t}$, and $f_{i, t-1} \in \mathbb{R}^{C_{Fi} \times H_{Fi} \times W_{Fi}}$, $s_{di}$, $s_{i, t}$, and $s_{i, t-1} \in \mathbb{R}^{C_{Si} \times H_{Si} \times W_{Si}}$.

\begin{figure}[t]
\centering
\includegraphics[width=8cm]{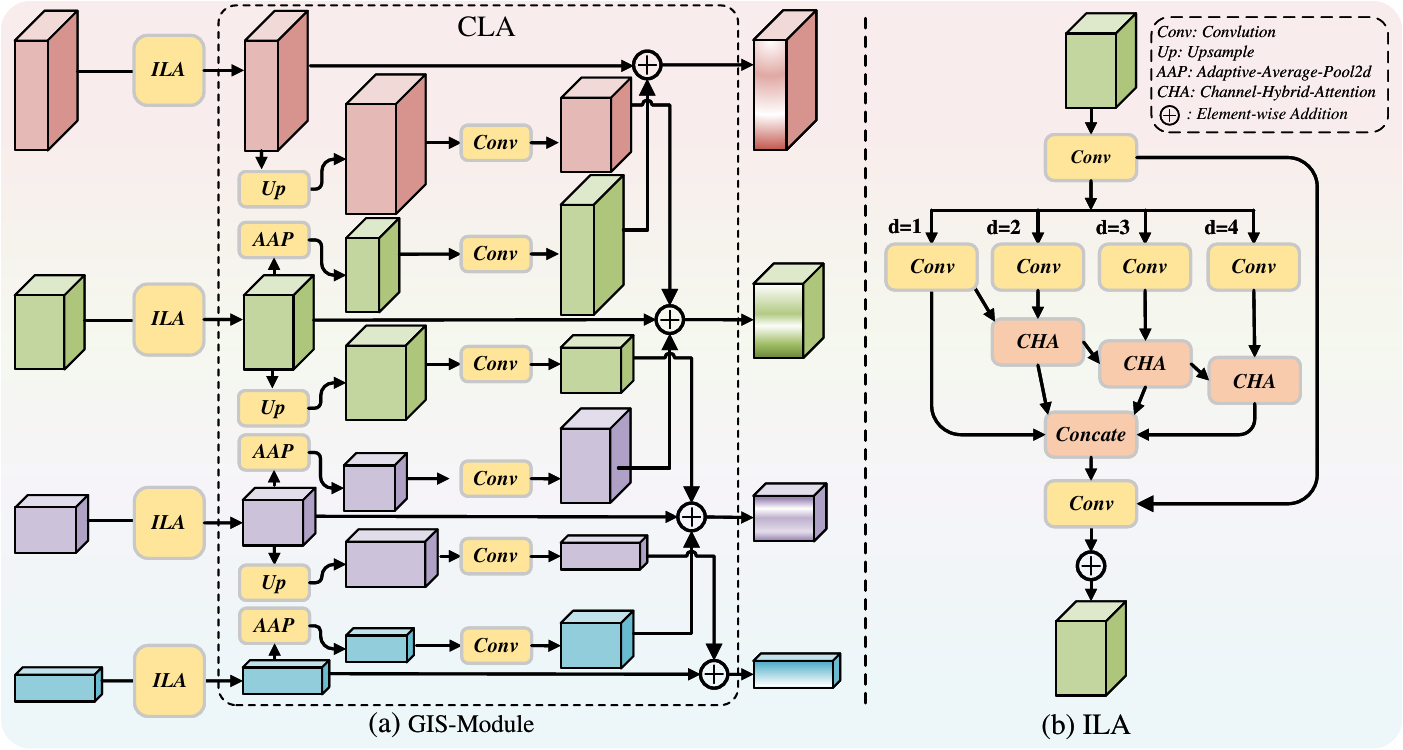}
\caption{The proposed GCS-Module comprises two processes: Intra-Level Multi-scale Feature Aggregation (ILA) and Cross-Level Feature Semantic Alignment (CLA). The ILA aggregates multi-scale information embedded within features at each layer using a Hierarchical Feature Fusion (HFF) structure connected by channel-mixing attention, as shown in (b). The CLA unit then aligns information across layers to establish a globally refined feature representation.}
\label{fig:skipconnection}
\end{figure}

\subsection{GCS-Module for Feature Refinement}

The GCS-Module is proposed to extract and aggregate multi-scale driver facial features to establish a globally refined feature representation. The structure of the GCS-Module is shown in \textbf{Fig. \ref{fig:skipconnection}(a)}, which comprises two processes: Intra-Level Multi-scale Feature Aggregation (ILA) and Cross-Level Feature Semantic Alignment (CLA).

To extract the multi-scale information embedded within features at each layer, we designed the ILA, with its main structure depicted in \textbf{Fig. \ref{fig:skipconnection}(b)}.

Firstly, the input features $f_{di} \in \mathbb{R}^{C_{Fi} \times H_{Fi} \times W_{Fi}}$, where $i \in \{1, 2, 3, 4\}$ , are mapped through group point convolutions to $k_i$ groups of features, $f_{ig, n} \in \mathbb{R}^{\frac{C_{Fi}}{k} \times H_{Fi} \times W_{Fi}}$, where $k_i = 2 + 2 \cdot \left\lfloor \frac{i-1}{2} \right\rfloor$, $n \in \{1, \ldots, k_i\}$, $\left\lfloor \cdot \right\rfloor$ denotes the floor function, which rounds down to the nearest integer. Then, $k$ dilated convolution branches with different dilation rates learn representations in parallel, resulting in $f_{igr, n} \in \mathbb{R}^{\frac{C_{Fi}}{k_i} \times H_{Fi} \times W_{Fi}}$. To mitigate grid artifacts potentially caused by large receptive fields, we employ a Hierarchical Feature Fusion (HFF) \cite{mehta2018espnet} structure to integrate the feature maps. Additionally, we use a Channel Integration Unit (CIU) that sequentially combines feature maps from adjacent branches, avoiding issues of information redundancy or dilution that simple addition might cause.

\begin{equation}
\begin{aligned}
f_{ifuse,n} &= CIU(f_{igr,n-1}, f_{igr,n}) \\
\end{aligned}
\end{equation}
where, $f_{ifuse,n} \in \mathbb{R}^{\frac{C_{Fi}}{k_i} \times H_{Fi} \times W_{Fi}}$ is the fused feature. 
The process of CIU is as:
\begin{equation}
C_{att}= \sigma (\omega_{i-1} \cdot F_{Avg} (f_{i-1}) + \omega_{ i } \cdot F_{ Avg } (f_{i}))
\end{equation}
\begin{equation}
\begin{aligned}
x_{ci} &= C_{att} \cdot Conv_{3 \times 3}(Cat[x_{i-1}, x_{i}]) + x_{i-1}
\end{aligned}
\end{equation}
where, $F_{Avg}(\cdot)$ is the average pooling, $\omega_{i}$ are the weights of channel features, $\sigma$ is the sigmoid function, and $C_{att} \in \mathbb{R}^{{C_{F(i-1)}} \times 1 \times 1} $ is the channel hybrid attention mask.
After layer-wise fusion, the results are mapped through point convolution and connected by residual connections to generate the integrated feature representation $f_{mi} \in \mathbb{R}^{C_{Fi} \times H_{Fi} \times W_{Fi}}$:
\begin{equation}
f_{mi}= f_{di}+Conv_{3 \times 3}(Cat[f_{ifr, 1}, f_{ifuse, 2} \dots , f_{ifuse, k_i}])
\end{equation}
Then, to amalgamate diverse semantic information across multiple feature layers and bolster consistent feature expression, we propose the CLA unit, which focuses on aggregating anchor feature maps from the base layer alongside associated feature sets from neighboring layers. Specifically, this unit takes $f_{mi}, i \in \{1, 2, 3, 4\}$ as an anchor feature map, and $N_i$ is a corresponding set of adjacent layer features:
\begin{equation}
N _ { i }  = \begin{cases} 
f_{m(i+1)}, &i = 1  \\
(f_{m(i-1) }, f_{m(i+1)}), & i \in \{2,3\}\\
f_{m(i-1)}, & i = 4 
\end{cases}
\end{equation}
Firstly, each feature in $N_i$ is adjusted to the same spatial dimension as $f_{mi}$ using adaptive average pooling or bilinear upsampling. Convolutional layers are then applied to transform features into a uniform feature space and integrate them with the anchor feature map $f_{mi}$ through element-wise addition to enable multi-level information interaction. $f_{mi}$ are processed by the above operations to get the output $f_{si}$. This semantic alignment process is denoted as $CLA$.
\begin{equation}
f_{si} = CLA(f_{mi})
\end{equation}
where, $i \in {1, 2, 3, 4}$.
At this point, the input driver facial features $f_{di} \in \mathbb {R}^{C_{Fi} \times H_{Fi} \times W_{Fi}}$ have been processed through the GCS-Module, resulting in the output $f_{si}, i \in {1, 2, 3, 4}$ and maintaining the same dimensions.

\begin{figure}[t]
\centering
\includegraphics[width=8cm]{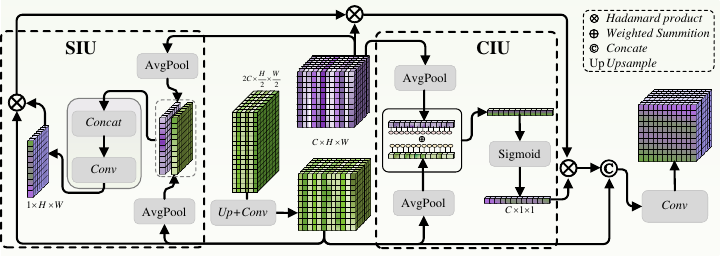}
\caption{
Channel-Spatial Hybrid Attention (CSHA). CSHA is proposed to recover lost detail information and enhance feature representation during decoding. CIU and SIU represent Channel Integration Unit and Spatial Integration Unit, respectively.}
\label{fig:up}
\end{figure}

\subsection{W-Net for Feature Alignment}

We propose a novel cross-domain architecture W-Net, which is universally applicable to the systematic integration of information from any two disparate domains or modalities. Its distinguishing characteristic resides in the ``Encoding-Independent Partial Decoding-Fusion Decoding" W-shaped framework, which employs skip connections to facilitate multi-scale feature integration throughout the encoding and decoding phases. The Two-Stage Decoding (TS-Decoding) strategy is the essence of W-Net, whose main idea is to perform intra-domain feature alignment before feature fusion decoding. This ensures that the features extracted from each domain are semantically consistent, thereby guaranteeing that the subsequent fusion process operates on well-aligned, high-quality feature representations.

\subsubsection{Independent Partial Decoding}
To address semantic inconsistencies in feature aggregation from driver's face and road scene images, we introduce independent partial decoding to decode each feature type separately before fusion, ensuring semantic alignment within each domain. Our Channel-Spatial Hybrid Attention (CSHA) is designed to better recover lost detail information and enhance feature representation, as illustrated in \textbf{Fig. \ref{fig:up}}.


\textbf{Channel-Spatial Hybrid Attention (CSHA)}: We input $x_i \in \mathbb {R}^{C_{Fi} \times H_{Fi} \times W_{Fi}}$ along with the shallower feature map $x_{i-1} \in \mathbb {R}^{C_{F(i-1)} \times H_{F(i-1)} \times W_{F(i-1)}}$ into CSHA. The dimension of $x_i$ is transformed to $\mathbb {R}^{C_{F(i-1)} \times H_{F(i-1)} \times W_{F(i-1)}}$ through upsampling operations and a linear transformation. Then, both $x_i$ and $x_{i-1}$ are fed into the CIU for channel integration to emphasize important cross-channel relationships:
\begin{equation}
x_{ci}=CIU(x_i,x_{i-1})
\end{equation}

To capture the spatial contextual relationship between the two features, we design the Spatial Integration Unit (SIU). The specific process is as:
\begin{equation}
S_{att} = \delta \left(Conv_{7 \times 7}\left(Cat\left(Mean(x_i), Mean(x_{i-1})\right)\right)\right)
\end{equation}
\begin{equation}
x_{csi} = x_{ci} \ast S_{att}
\end{equation}
where, $Mean(\cdot)$ indicates calculating the mean across the channel dimensions, and $\ast$ indicates the Hadamard product. $S_{att} \in \mathbb {R}^{1 \times H_{F(i-1)} \times W_{F(i-1)}}$ represents spatial integration attention, and $x_{csi} \in \mathbb {R}^{{C_{F(i-1)} \times H_{F(i-1)} \times W_{F(i-1)}}}$ denotes the output of CSHA unit.

For facial features, we independently decode $f_{si}, i \in {1, 2, 3, 4}$, and by cascading three CSHA units, we obtain the corresponding output features $f_{pi}$, $i \in {1, 2, 3, 4}$.

\begin{equation}
f_{pi} = \begin{cases}
CSHA(f_{p(i+1)}, f_{si}), &i \in {(1,2,3)}\\
f_{si}, &i=4
\end{cases}
\end{equation}
Similarly, the driving scene features $s_{di}$ are independently decoded to obtain the output features$s_{pi}$, where $i \in {1, 2, 3, 4}$.
\begin{equation}
s_{pi} = \begin{cases}
CSHA(s_{d(i+1)}, s_{di}), &i \in {(1, 2, 3)}\\
s_{di}, &i=4
\end{cases}
\end{equation}

\subsubsection{Fusion Decoding}
To deeply integrate driver facial and scene features, we leverage the capability of multi-head cross-attention (MHCA) \cite{vaswani2017attention} to capture long-range dependencies for feature fusion decoding.

\textbf{Multi-Head Cross-Attention (MHCA)}: The MHCA unit mainly consists of three components: feature encoding, cross-attention (CA), and a feed-forward network (FFN). The unit takes the driving scene features $s_{pi} \in \mathbb{R}^{C_{Si} \times H_{Si} \times W_{Si}}$ and the driver facial features $f_{pi} \in \mathbb{R}^{C_{Fi} \times H_{Fi} \times W_{Fi}}$ as input, where $i \in {1, 2, 3, 4}$, $C_{Si} = C_{Fi} = 64 \times 2^{(i-1)}$.
First, encode the two types of input features separately:
\begin{equation}
 f_{ei} = Flatten(Conv_{fi \times fi }(f_{pi})+Pos_{fi})  
\end{equation}  
\begin{equation}
 s_{ei} = Flatten(Conv_{si \times si }(s_{pi})+Pos_{si})  
\end{equation}
where, $Pos_{fi}$ is the corresponding position embedding of $f_{ei}$, $Conv_{fi \times fi}$ is the convolution operation with both the kernel size and stride being $fi$, and $Flatten$ is the flattening operation. The same applies to the operations on scene features.

Scene and facial features are processed through linear transformations and divided into multiple subsequences. Attention is computed within different subspaces, and the results are then combined,
\begin{equation}
    Head_{n} = CA(s_{ei}, f_{ei})
\end{equation}
\begin{equation}
   out = Linear(Cat(Head_1, \dots, Head_n))
\end{equation}
where, $n$ is the number of heads, and $d_k$ is the dimension of each subspace. For each head, the linearly transformed features $s_{ei}$ serve as the Query (Q), while the transformations of $f_{ei}$ serve as Key (K) and Value (V). FFN integrates the outputs from all attention heads using a Linear layer. Finally, the integrated features are reconstructed back to the original spatial dimensions of $s_{ei}$.

After layer-by-layer fusion of the scene features and driver features, the fused features $d_{i},i \in {1, 2, 3, 4}$ are obtained.

\subsubsection{Decoding Head}
We perform three-step sequential decoding on $d_{i}$ using the CSHA unit, where the final step yields the shape fusion feature $d_u \in \mathbb{R}^{64 \times \frac{H_S}{4} \times \frac{W_S}{4}}$. Then, through the head, we obtain an attention heatmap with the same resolution as the input scene image. The process can be shown as:
\begin{equation}
    ECA(d_u) = d_u \cdot (\sigma(Conv_{3 \times 3}(F_{avg}(d_u))))
\end{equation}
\begin{equation}
    d_{hr} = ConvT(ECA(d_u))
\end{equation}
\begin{equation}
\begin{aligned}
AC(d_{hr}) &= Conv_{3 \times 3} \left( Cat\left[ Conv_{1 \times 1}(d_{hr}), \right. \right. \\
&\phantom{{}= Conv_{3 \times 3} \left( Cat\left[\right.\right.} \left. \left. Conv_{3 \times 3}(d_{hr}) \right] \right)
\end{aligned}
\end{equation}
\begin{equation}
    HeatMap = \sigma(AC(d_{hr}))
\end{equation}
where, $d_{hr} \in \mathbb{R}^{16 \times H_S \times W_S}$, $HeatMap \in \mathbb{R}^{1 \times H_S \times W_S}$ is the final predicted driver attention map, which has the same resolution as the input driving scene. $ConvT$ refers to two consecutive deconvolution layers, $AC(\cdot)$ denotes a channel adjustment unit.

\section{Experiments}

In this section, we provide a detailed description of the dataset, evaluation protocols, loss function, ground truth, and implementation details.

\subsection{Experimental Setup}

\subsubsection{Dataset}
The Look Both Ways (LBW) \cite{kasahara2022look} dataset is the only publicly available dataset containing driver attention-scene associations for driving scene videos, driver face videos, and driver 3D gaze ground truth values. Consequently, our work is primarily based on the LBW dataset. It Comprises 6.8 hours of free driving on public roads and includes a total of 123,297 synchronized driver face and stereoscopic scene images with 3D gaze ground truth values, captured with the participation of 28 drivers (22 males and 6 females). 

\subsubsection{Evaluation Protocols}

The evaluation protocol includes distribution-based, consistency-based, and position-based metrics. Distribution-based metrics encompass Kullback-Leibler divergence (KL) and similarity metric (SIM), which measure the difference and similarity between two probability distributions, respectively. Consistency-based metrics include the correlation coefficient (CC), which quantifies the linear correlation between predicted and actual values. Position-based metrics include normalized scanpath saliency (NSS), which assesses the overlap between predicted attention regions and actual  regions. Lower KL values and higher SIM, CC, and NSS values indicate better performance. In our paper, the symbols used in all tables are defined as follows: $\downarrow$ indicates that smaller values are better. $\uparrow$ indicates that larger values are better. Best values are marked in bold.

\subsubsection{Loss Functions}

We evaluate the model comprehensively using a combination of metrics, including Normalized Scanpath Saliency (NSS), Kullback-Leibler divergence (KL), Correlation Coefficient (CC), and Similarity Index (SIM). We optimize the parameters using a loss function composed of four metrics during the training of the EraW-Net network.
\begin{equation}
\begin{aligned}
        L(E,G)=&\rho_1 NSS(E,G)+\rho_2KL(E,G)\\
            &+\rho_3SIM(E,G)+\rho_4CC(E,G)
\end{aligned}
\end{equation}

where, $E$ is the network's estimated driver attention map; $G$ is the ground ground truth attention map; $\rho$ is the weight of each loss (empirically set as $\rho_1=-0.005,\rho_2=1,\rho_3=-0.2,\rho_4=-0.1$; $NSS$, $KL$, $SIM$, and $CC$ represent the NSS loss, KL loss, SIM loss, and CC loss, 
\begin{equation}
    NSS(E,G)=\frac{1}{N} \ast \sum_i\frac{E_{i}-\mu(E)}{\sigma(E)} 
\end{equation}
\begin{equation}
    KL(E,G)=\sum_{i}G_{i}\ast \log(\frac{E_{i}}{E_{i}+\delta}+\delta)
\end{equation}
\begin{equation}
    SIM(E,G)=\sum_{i}\min(\frac{E_{i}}{\sum_{i=1}^{n}E_{i}+\delta},\frac{G_{i}}{\sum_{i=1}^{n}G_{i}+\delta}) 
\end{equation}
\begin{equation}
    CC(E,G)=\frac{Cov(E,G)}{\sigma(E)\ast\sigma(G)}
\end{equation}
where, $E_i$ and $G_i$ represent the points in $E$ and $G$, $\delta$ is a regularization constant to prevent division by zero, $Cov(\cdot)$ denotes the covariance function, and $\sigma$and $\mu$ are the standard deviation and mean.

\subsubsection{Ground Truth of Attention Maps} We conduct experiments using two types of ground truth maps: the gaze-projected heatmap and the sequential fixation heatmap.

\emph{Gaze-Projected Heatmap:} In the LBW dataset, the ground truth of driver attention visual saliency $s_g(x)$ is modeled as a function of the 3D gaze direction $g$, predicted from facial appearance images $I_g$. Given the depth estimation from stereo scene cameras in $X$, 3D points are reconstructed and projected onto the center of the eye $e$ to form direction $s$. The angular difference between $s$ and gaze direction $g$ is used to model the projected scene saliency $s_g$.

\emph{Sequential Fixation Heatmap:}
During the generation of gaze-projected heatmaps, we Observed that inaccuracies in the attention maps were caused by missing parts of the depth maps. To improve this issue and and account for the continuity of the driver’s gaze, we adopted a method similar to that used in the DR(eye)VE dataset \cite{palazzi2018predicting} to create a ``sequential fixation heatmap" over a period of 1 second.
We selected multiple frames within a temporal sliding window, and mapped the gaze points from each frame to the coordinates of the current frame using homography transformation. We modeled the probability distribution of gaze points using a Gaussian function with fixed parameters and introduced a spatio-temporal weighting mechanism that assigns weights based on the proximity of gaze points in both space and time. By integrating weighted Gaussian distributions, we generated the sequential fixation heatmaps representing the driver’s visual attention. The weight calculation process can be described as:

\begin{equation}
\Delta x_k = H_k(p{k_x}) - p{c_x}
\end{equation}
\begin{equation}
\Delta y_k = H_k(p{k_y}) - p{c_y}
\end{equation}
\begin{equation}
\Delta t_k = \frac{|k - c|}{a}
\end{equation}
\begin{equation}
d_k = \sqrt{(\Delta x_k)^2 + (\Delta y_k)^2 + (\Delta t_k)^2}
\end{equation}
\begin{equation}
    w_k = e^{-\frac{d_k^2}{2\cdot \sigma^2}}
\end{equation}

where, $p{c}$ is the gaze point in the current frame, $p{k}$ is the gaze point in frame $k$, $H_k (\cdot)$ represents the homography transformation, $a \in \mathbb{R}$ is an adjustable real parameter. $d_k$ denotes the total distance, $w_k$ is the corresponding weight, and $\sigma=200$ is the standard deviation of the Gaussian function.

\subsubsection{Implementation Details}

During training, we use AdamW as the optimizer and CyclicLR \cite{smith2017cyclical} as the learning rate scheduler, with learning rate boundaries from 2e-6 to 1e-5. Our EraW-Net is implemented in a Python 3.7.2 and Pytorch 1.12.1 environment, equipped with 4 NVIDIA GeForce RTX 4090 GPUs and an Intel(R) Xeon(R) Gold 6326 CPU @ 2.90GHz. We employ mixed-precision training to balance speed and accuracy, with a batch size of 32.

\subsection{Comparison Experiments}

\subsubsection{Comparison with State-of-the-Art Methods}

To the best of our knowledge, there are currently no existing works that perform end-to-end (E2E) pixel-level driver attention estimation correlated with driving scenes similar to ours. Therefore, we generate 3D gaze estimation results using state-of-the-art 3D gaze estimation methods and subsequently produce the corresponding attention prediction results using the gaze-projected heatmap generation method described in Section IV-A. We evaluate these results against the ground truth to compute the KL, CC, SIM, and NSS metric values. We compare our model against several state-of-the-art driver attention prediction methods, including XGaze \cite{zhang2020eth}, Full-Face \cite{zhang2017s}, GazeTR \cite{cheng2022gaze}, L2CS \cite{abdelrahman2023l2cs}, PureGaze \cite{cheng2022puregaze}, GazePTR \cite{cheng2024you}, and LookBothWays \cite{kasahara2022look}. Experiments are conducted on the LBW dataset. Notably, only our method and the LookBothWays integrate information from both the driver's face and the driving scene, while the other methods rely solely on driver facial image information.

\begin{table}[t]
\caption{COMPARATIVE EXPERIMENTS EMPLOYING GAZE-PROJECTED HEATMAPS AS GROUND TRUTH.}
\label{tab:comparion s_g} 
\centering
\renewcommand{\arraystretch}{1.2}
\renewcommand\tabcolsep{5.0pt}
\begin{tabular}{@{}lcccccc@{}} %
\toprule
Method&Data & E2E  & KL$\downarrow$ & CC$\uparrow$   &SIM$\uparrow$  &NSS$\uparrow$ \\ \midrule
XGaze \cite{zhang2020eth} & F &\ding{55} &1.4134	&0.5934	 &0.5203  &2.3731  \\
Full-Face \cite{zhang2017s} & F &\ding{55} & 1.2065  & 0.6256  & 0.5447  & 2.5015  \\
GazeTR \cite{cheng2022gaze}& F &\ding{55} & 1.3800  & 0.6133  & 0.5363  &2.4440  \\
L2CS \cite{abdelrahman2023l2cs}& F &\ding{55} &1.2661  & 0.6250  & 0.5448  & 2.4938 \\
PureGaze \cite{cheng2022puregaze}& F &\ding{55} & 1.1978  &0.6252	&0.5413  &2.4986 \\
GazePTR \cite{cheng2024you} & F &\ding{55} & 1.2362	&0.6302	 &0.5499  &2.5211	\\
LookBothWays \cite{kasahara2022look}& F+S &\ding{55} & 1.1413  & 0.6406	 & \textbf{0.5555}  & 2.5582  \\
EraW-Net(Ours)&F+S &\ding{51} & \textbf{0.9299} & \textbf{0.6624} & 0.5485 & \textbf{2.6273} \\ \bottomrule
\end{tabular}
\end{table}

\begin{table}[t]
\caption{COMPARATIVE EXPERIMENTS EMPLOYING SEQUENTIAL FIAXTION HEATMAPS AS GROUND TRUTH.}
\label{tab:comparion fixmap} 
\centering
\renewcommand{\arraystretch}{1.2}
\renewcommand\tabcolsep{5.0pt}
\begin{tabular}{@{}lcccccc@{}} %
\toprule
Method &Data & E2E & KL$\downarrow$ & CC$\uparrow$   &SIM$\uparrow$  &NSS$\uparrow$  \\ \midrule
XGaze \cite{zhang2020eth}& F  & \ding{55}  & 3.2240  & 0.3174  & 0.2409	  & 2.5007  \\
Full-Face \cite{zhang2017s}& F & \ding{55} & 3.0257  & 0.3336	& 0.2528  & 2.6297  \\
GazeTR \cite{cheng2022gaze}& F & \ding{55} & 3.1054  & 0.3469  & 0.2570  & 2.7370 \\
L2CS \cite{abdelrahman2023l2cs}& F & \ding{55} & 3.0679  & 0.3331  & 0.2499  & 2.6296 \\
PureGaze \cite{cheng2022puregaze}& F & \ding{55} & 2.9636  & 0.3281  & 0.2558	&2.5828	\\
GazePTR \cite{cheng2024you}& F  & \ding{55}  & 3.0281  &0.3467  & \textbf{0.2581}  &2.7324	\\
LookBothWays \cite{kasahara2022look}& F+S & \ding{55} & 3.0160  & 0.3334 & 0.2517 & 2.6265  \\
EraW-Net(Ours)& F+S  & \ding{51}  & \textbf{2.4054}  & \textbf{0.3496}	& 0.2422  & \textbf{2.7883} \\ \bottomrule
\end{tabular}
\end{table}

\textbf{Tables \ref{tab:comparion s_g}} and \textbf{\ref{tab:comparion fixmap}} present the experimental results using the gaze-projected heatmaps and the sequential fixation heatmaps as ground truth, respectively. Rapid and frequent changes in driver attention in real driving scenes lead to generally poor performance across all methods when the sequential fixation map is used as the ground truth. F represents the input face image and S represents the corresponding scene image. It is evident that methods using both driver facial images and driving scenes as inputs outperform those using only driver facial images. This demonstrates that driving scenes indeed provide valuable information for estimating driver attention. In \textbf{Table \ref{tab:comparion s_g}} and \textbf{Table \ref{tab:comparion fixmap}}, our model achieves the best performance in the KL, CC and NSS metrics, with the KL metric outperforming the second-best result by 18.5\% and 20.2\%, respectively. This improvement can be attributed to our proposed W-Net architecture, which effectively integrates cross-view information from both the driver's facial features and the driving scene. Additionally, The introduced DAF-Module captures critical dynamic cues, enhancing the model's adaptability to dynamic scenes in real-world driving environments. Meanwhile, the GCS-Module refines the driver's facial features, enabling the model to accurately capture the driver's state. However, as observed in both tables, our model generally underperforms in the SIM metric. 

Analysis of underperforming metric: By incorporating visualized results into our analysis \textbf{(refer to Supplementary Material, Fig. S1)}, we have identified the reasons why EraW-Net performs relatively poorly on the SIM metric. We attribute this discrepancy to two primary factors: (1) Among all the methods compared, only our model employs an end-to-end prediction approach. This method requires not only predicting correct attention focus locations but also ensuring that the predicted attention distribution closely aligns with the ground truth. In contrast, other methods generate attention maps based on predicted 3D gaze vectors combined with depth information and camera calibration information, naturally resulting in attention distributions similar to the ground truth maps. This inherent advantage particularly benefits these methods in metrics like SIM and NSS, posing a significant challenge for our approach. (2) Our predictions show irregular attention distributions, reflecting the driver's scanning behavior, notably evident when using sequential fixation heatmaps as ground truth. While our distribution differs from the ground truth, which focuses near a single point, it accurately represents actual driver scanning patterns. This discrepancy aligns with realistic driver saccadic movements but contrasts with the single-point focus of the ground truth. We believe that EraW-Net's performance on the SIM metric is affected due to these reasons.


\begin{figure}[t]
\centering
\includegraphics[width=8cm]{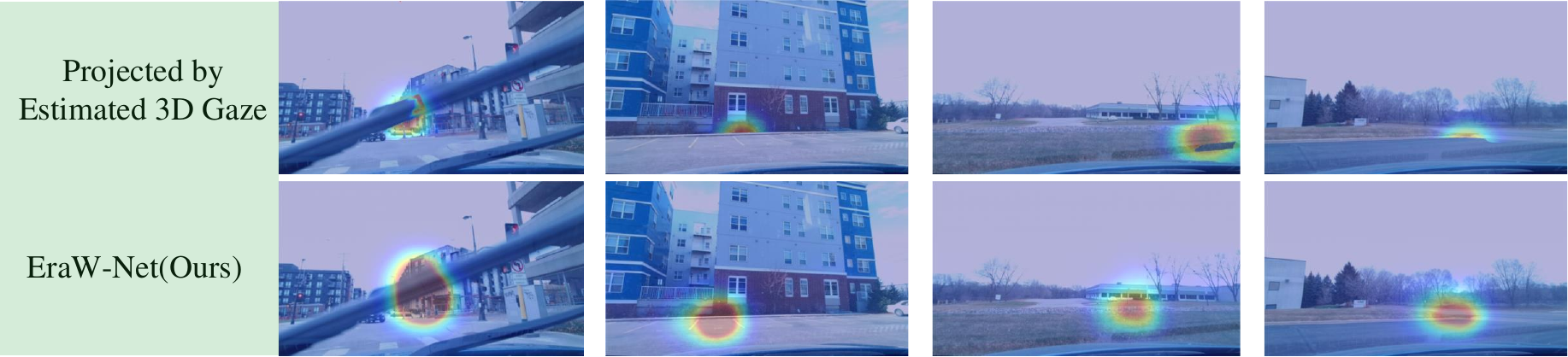}
\caption{The attention maps in the first row are generated by state-of-the-art 3D gaze estimation methods, whose low quality are due to depth inaccuracies. The second row shows the corresponding attention maps generated by our end-to-end model, EraW-Net.}
\label{fig: depth errors}
\end{figure}

As we mentioned earlier, we are the only end-to-end model in these methods. The methods of mapping the driver's 3D gaze information to the driving scene requires additional information such as scene depth information and camera parameters, whose accuracy is closely related to the quality of depth estimation; inaccurate depth estimation can lead to inaccurate attention mapping results, as shown in the \textbf{Fig. \ref{fig: depth errors}}. In comparison, our end-to-end estimation has the following advantages: Firstly, Our model does not rely on depth estimation information, thereby reducing uncertainties associated with depth estimation errors and enhancing the stability and reliability of the system. Additionally, requires only images from both inside and outside the vehicle, eliminating the need for complex camera calibration processes and simplifying the data processing pipeline.  Moreover, our method requires input from just two cameras for end-to-end attention-scene correlation. This simplification reduces system design and data processing complexity, thereby lowering deployment costs. This simplification of system setup and usage, along with increased applicability and robustness, allows our model to be more broadly applied in various driving environments.

\subsubsection{Comparison Experiments of Information Integration}

\begin{figure}[t]
\centering
\includegraphics[width=8cm]{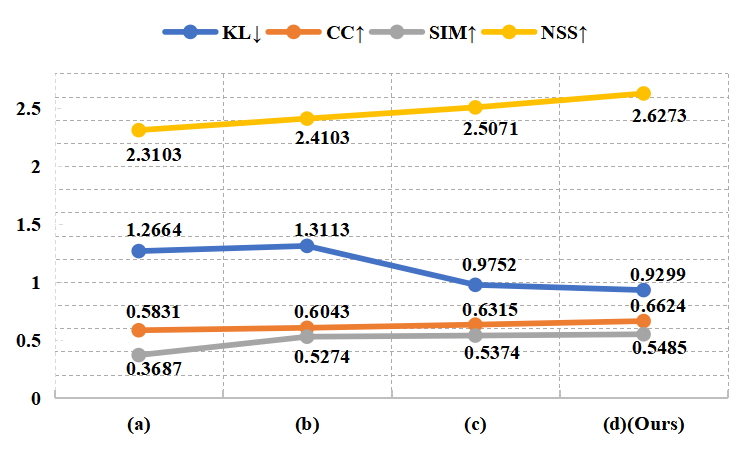}
\caption{Ablation study on different information integration strategies.}
\label{fig:fusion stragegy exp}
\end{figure}

In \textbf{Fig. \ref{fig:fusion classification}}, we present three classic information integration strategies: (a) Decision-Level Fusion, (b) Late-Feature-Level Fusion, and (c) Hierarchical-Feature-Level Fusion. \textbf{Fig. \ref{fig:fusion classification}(d)} shows our proposed W-Net architecture for cross-domain integration. We conducted experiments based on these four integration methods using the gaze-projected heatmap as ground truth, and the results are summarized in \textbf{Fig. \ref{fig:fusion stragegy exp}}. It is evident that our W-Net outperforms all other methods across all evaluated metrics.

For the three fusion strategies (a), (b), and (c), we observe that both the late-feature-level fusion method (b) and the hierarchical feature-level fusion method (c) significantly outperform the decision-level fusion (a) across all metrics except for the KL divergence metric. Notably, method (c) achieves an average of 21.4\% better overall performance compared to method (a). This indicates that feature-level fusion methods are more effective in leveraging dual information sources from the driver. Comparing the results of the feature-level fusion strategies (b) and (c), the hierarchical feature-level fusion method (c) demonstrates substantial improvements over the late-feature-level fusion method (b), particularly with enhancements exceeding 4\% in the KL, CC, and NSS metrics. These findings suggest that hierarchical feature-level fusion integrates and utilizes the input data sources more effectively, thereby enhancing overall model performance.

Further analysis of the experimental results for our W-Net and the aforementioned fusion methods reveals that the W-Net consistently outperforms strategies (a), (b), and (c) across all metrics. Notably, it demonstrates a 4.6\% decrease in the KL metric and a 4.8\% increase in the NSS metric compared to method (c). Our proposed W-Net exhibits superior performance across all metrics when compared to fusion-based strategies, validating its strong advantages in complex information integration tasks. This superior performance is attributed to our ``Encoding-Independent Partial Decoding-Fusion Decoding" W-Net architecture. The two-stage decoding strategy effectively addresses the limitations of unimodal feature extraction and semantic inconsistencies, leveraging driving scene information to complement driver facial information and enhancing the accuracy of driver attention estimation.



\subsubsection{Comparison Experiments of the Fusion Modules}

There are several methods to fuse the two features. At the fusion decoding stage, we employed Multi-Head Cross-Attention (MHCA) to integrate the feature representations from the driver’s face and the driving scene. To identify the optimal fusion method, we conducted comparative experiments. Specifically, we first aligned the two features into the same dimension and then utilized four fusion modules, namely ADD, AFF \cite{dai2021attentional}, SE \cite{hu2018squeeze}, and Conv, to replace MHCA for fusing the two features.

ADD denotes direct fusion of the two feature maps through addition. For Conv, feature maps are concatenated and then processed through convolution. The SE module applies squeeze-and-excitation to concatenated features for channel-wise attention, while AFF module utilizes the Attentional Feature Fusion module to jointly capture spatial-temporal and channel-specific attention. The experimental results are summarized in \textbf{Table \ref{tab:fusion modules s_g}}.

We conducted two sets of ablation experiments using the
gaze-projected heatmap as ground truth. The experimental results indicate that the MHCA fusion method we employed achieved optimal performance across all metrics, surpassing the second-best results in the KL, CC, SIM, and NSS metrics by 12.6\%, 7.0\%, 5.3\%, and 5.8\%, respectively, indicating the superior advantage of MHCA in integrating the two features. This underscores its ability to effectively capture the intricate relationship between driver facial features and driving scene characteristics, thereby enhancing the overall performance of the model.


\subsection{Ablation Study}

\begin{table}[t]
\caption{ABLATION STUDY ON DIFFERENT FUSION MODULES.}

\label{tab:fusion modules s_g} 
\centering
\renewcommand{\arraystretch}{1.2}
\renewcommand\tabcolsep{5.0pt}
\begin{tabular}{@{}lcccc@{}} %
\toprule
Fusion& KL$\downarrow$ & CC$\uparrow$   &SIM$\uparrow$  &NSS$\uparrow$ \\ \midrule
ADD    & 1.0982	& 0.6172	& 0.5122	& 2.4506  \\
AFF      & 1.1060	& 0.6258	& 0.5050	& 2.4839 \\
Conv    & 1.0636	& 0.6082	& 0.5208	&2.4108 \\
SE    & 1.0989	& 0.6193	& 0.5133	& 2.4553    \\
MHCA (Ours)   & \textbf{0.9299} & \textbf{0.6624} & \textbf{0.5485} & \textbf{2.6273} \\ \bottomrule
\end{tabular}

\end{table}

\begin{table}[t]
\caption{ABLATION STUDY ON VARIOUS LOSS FUNCTIONS.}
\label{tab:loss function} 
\centering
\renewcommand{\arraystretch}{1.2}
\renewcommand\tabcolsep{5.0pt}
\begin{tabular}{@{}lcccc@{}} %
\toprule
Loss function  & KL$\downarrow$ & CC$\uparrow$   & SIM $\uparrow$  & NSS $\uparrow$ \\ \midrule
$CC$    &2.4247  &0.6377  &0.3085  &2.5276 \\
$NSS$   &2.4415  &0.6384  &0.4049   &2.5327 \\
$SIM$	&2.4315	 &0.6288  &0.5377	&2.4978  \\
$SIM+NSS$   &2.3681 &0.6237  & 0.5347   &2.4713 \\
$KL+NSS$   & 1.3793 & 0.6205 &0.3848 & 2.4630 \\
$CC+SIM+KL$   & 1.0300 & 0.6235 &0.5120 &2.4728\\ 
$CC+SIM+KL+NSS$  & \textbf{0.9299} & \textbf{0.6624} & \textbf{0.5485} & \textbf{2.6273} \\ \bottomrule
\end{tabular}

\end{table}

\subsubsection{Ablation Experiments of Loss Functions}

We investigated the performance implications of different loss functions using the gaze-projected heatmap as ground truth, summarized in \textbf{Table \ref{tab:loss function}}. By experimenting with different combinations of CC, NSS, KL, and SIM as loss functions, we observed that when CC, NSS, or SIM are used individually, the model's performance is relatively similar and generally suboptimal. However, combining loss functions, such as SIM+NSS, KL+NSS, and CC+SIM+KL, significantly optimizes results. Notably, the addition of the KL term markedly improves performance on the KL metric. Furthermore, using the combination of CC+SIM+KL+NSS as the loss function yields the best performance across all metrics. This is because the combination of these four metrics can effectively capture different aspects of the prediction accuracy, providing a more comprehensive evaluation and optimization of the model's performance.

\subsubsection{Ablation Experiments of Modules}

We designed ablation experiments to analyze the impact of the modules in our proposed EraW-Net on the LBW dataset. We constructed a U-Baseline model for estimating driver attention. It encodes both the driving scene images and the driver's facial images, and hierarchically decodes through the \emph{Fusion Decoding} and \emph{Decoding Head} described in part III. Method, showing a U-shape.

\begin{table}[t]
\caption{ABLATION EXPERIMENTS ON THE GAZE-PROJECTED HEATMAPS.}
\label{tab:ablation s_g} 
\centering
\renewcommand{\arraystretch}{1.2}
\renewcommand\tabcolsep{5.0pt}
\begin{tabular}{@{}lcccc@{}} %
\toprule
Method    & KL$\downarrow$ & CC$\uparrow$   & SIM $\uparrow$  & NSS$\uparrow$ \\ \midrule
U-Baseline  & 1.6755  & 0.5857 & 0.4303  & 2.3195         \\
U-Baseline+DAF   &1.1656	&0.5928	&0.4837	&2.3532       \\
U-Baseline+DAF+GCS	&0.9752	&0.6315	&0.5374	&2.5071         \\
W-Net      & 1.1007 & 0.5974  & 0.5089  & 2.3646        \\
W-Net+DAF 	&0.9774	&0.6287	&0.5358	&2.4971         \\
W-Net+DAF+GCS  & \textbf{0.9299} & \textbf{0.6624} & \textbf{0.5485} & \textbf{2.6273}     \\ \bottomrule
\end{tabular}


\end{table}

\begin{table}[t]
\caption{ABLATION EXPERIMENTS ON THE SEQUENTIAL FIXATION HEATMAPS.}
\label{tab:ablation fixmap} 
\centering
\renewcommand{\arraystretch}{1.2}
\renewcommand\tabcolsep{5.0pt}

\begin{tabular}{@{}lcccc@{}} %
\toprule
Method    & KL$\downarrow$ & CC$\uparrow$   & SIM $\uparrow$  & NSS$\uparrow$ \\ \midrule
U-Baseline   &3.2351	&0.3172	&0.1993	&2.4939   \\
U-Baseline+DAF  &2.5345	&0.3452	&0.2345	&2.7400 \\
U-Baseline+DAF+GCS  &2.5160	&0.3481	&0.2237	&2.7791 \\
W-Net    & 2.8510 &0.3297 &0.2060 &2.6164    \\
W-Net+DAF  & 2.4273 & 0.3495 &0.2394 & 2.7872        \\
W-Net+DAF+GCS  & \textbf{2.4054}& \textbf{0.3496} &\textbf{0.2422} &\textbf{2.7883}         \\ \bottomrule
\end{tabular}


\end{table}

\begin{figure*}[t]
\centering
\includegraphics[width=16cm]{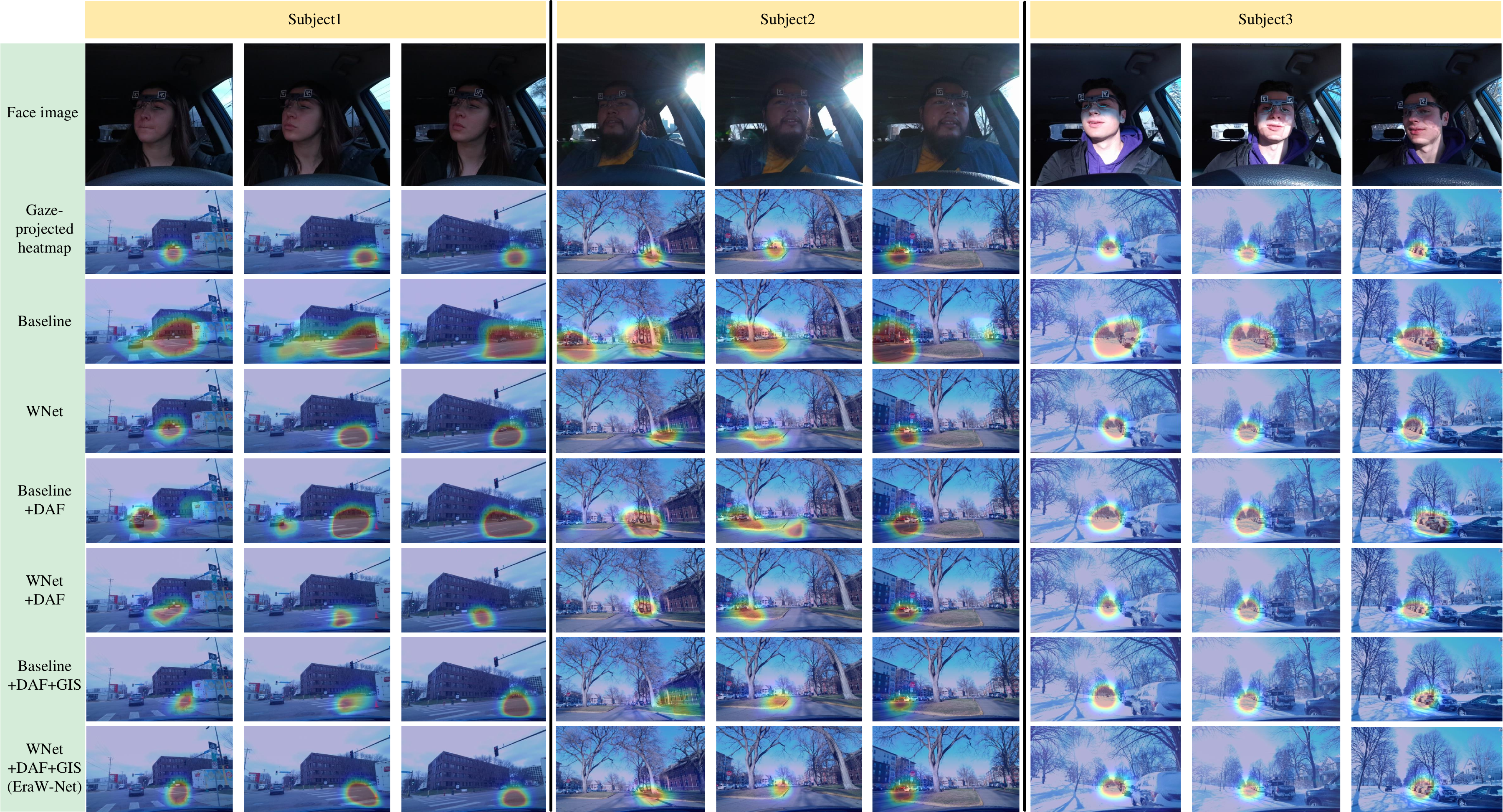}
\caption{Example results of our EraW-Net with different ablation versions using gaze-projected heatmap as ground truth on LBW dataset.}
\label{fig:s_g}
\end{figure*}


We conducted two sets of ablation experiments using the gaze-projected heatmap and the sequential fixation heatmap as ground truth, as shown in \textbf{Tables \ref{tab:ablation s_g}} and \textbf{\ref{tab:ablation fixmap}}, respectively.

For the experiments utilizing the gaze-projected heatmap as the reference, as illustrated in \textbf{Table \ref{tab:ablation s_g}}, the U-Baseline exhibited values of 1.6755 for KL, 0.5857 for CC, 0.4303 for SIM, and 2.3195 for NSS in estimating driver attention. By incorporating the proposed W-Net architecture to enhance the integration and optimization of complementary features from the driver’s facial image and the driving scene, we observed significant performance improvements across all metrics. Specifically, the KL divergence decreased by 34.3\%, while the SIM metric increased by 18.3\%. These results underscore the significant advantages of the proposed W-Net framework in effectively accommodating two distinct data sources.

Moreover, we enriched the W-Net with the DAF-Module to bolster dynamic feature representation and enhance the model's adaptability to temporal changes. Upon integrating the DAF-Module into both the U-Baseline and W-Net, substantial improvements were observed across all metrics. Specifically, relative to the U-Baseline, the U-Baseline+DAF demonstrated a 30.4\% reduction in the KL metric and a significant 12.4\% improvement in the SIM metric. Similarly, compared to W-Net, the W-Net+DAF showed an 11.2\% decrease in KL and enhancements of 5.0\%, 5.3\%, and 5.6\% in CC, SIM, and NSS, respectively. These outcomes affirm the efficacy of the DAF-Module in effectively capturing critical dynamics.

We further enhanced the models by integrating the GCS-Module, resulting in noticeable improvements across all metrics. Particularly noteworthy was the U-Baseline+DAF+GCS configuration, which showed enhancements exceeding 6.5\% across all metrics, with significant gains in the CC and NSS metrics of 20.6\% and 16.3\%, respectively. A comparative analysis between the W-Net+DAF+GCS and W-Net+DAF configurations revealed substantial improvements in each metric, surpassing the performance of all other setups. These findings underscore the effectiveness of the proposed GCS-Module in utilizing multi-scale facial features, leading to improved accuracy in driver attention estimation.

A comparison of experiments incorporating identical modules into both the U-Baseline and W-Net architectures reveals that setups based on the W-Net consistently outperform their U-Baseline counterparts. This robustly demonstrates the superior capability of the proposed W-Net architecture in effectively modeling the complex relationship between the two input sources.


Qualitative results of different ablation versions of our model are presented in \textbf{Fig. \ref{fig:s_g}}. We selected three sets of driver images (Subject1, Subject2, and Subject3) engaged in right turn, left turn, and straight driving tasks, respectively. The first row of facial images shows that all drivers exhibit large-scale observational movements, making these samples difficult to estimate. The second row highlights substantial shifts in their attention, posing a significant challenge to the model's ability to capture attention. Comparing predictions across all versions, it is evident that the W-Net architecture outperforms the U-Baseline. The incorporation of the DAF-Module enhances sensitivity to crucial motion features during driving, while the GCS-Module better models the driver's attention state, particularly during complex observational tasks. Consequently, our final model (EraW-Net) demonstrates fewer redundant areas and more accurate attention estimation compared to other versions. The resulting heatmaps, correlated with road context, closely resemble real-world conditions. For the experiments using the sequential fixation heatmap as the ground truth, the results \textbf{(refer to Supplementary Material, Fig. S2)} validate the effectiveness of the proposed DAF-Module, GCS-Module, and W-Net architecture, similar to the analysis of \textbf{Table \ref{tab:ablation s_g}}. 

Overall, our proposed EraW-Net method demonstrates exceptional performance on the LBW dataset for end-to-end scene-associated driver attention estimation. This success is attributed to several critical factors. Firstly, the W-Net architecture integrates driver facial and driving scene information using an ``Encoding-Independent Partial Decoding-Fusion Decoding" structure, effectively modeling complex relationships across domains. Secondly, the DAF-Module captures crucial dynamic features, enhancing adaptation to the intricate dynamics of traffic scenes and driver attention shifts. Finally, the GCS-Module refines the feature representation of driver facial features. The effectiveness of these components is validated through ablation studies and analyses presented in \textbf{Tables \ref{tab:ablation s_g}} and \textbf{\ref{tab:ablation fixmap}}.

\section{Conclusion}
In this paper, we propose EraW-Net, a novel method for end-to-end association of driver attention with dynamic scenes, which exploits the implicit connection between these two views and achieves systematic cross-view integration. EraW-Net enhances critical dynamic visual cues, refines facial feature representation, and achieves semantic alignment in cross-view association. The Dynamic Adaptive Filter Module (DAF-Module) filters the redundant frequency components and highlights vital motion through joint frequency-spatial filtering, adapting to frequent and unpredictable driving dynamics. Additionally, the Global Context Sharing Module (GCS-Module) improves the model’s ability to track driver states amidst non-fixed facial poses by capturing hierarchical features from local details to overall structures. Moreover, the W-Net architecture is designed for two parallel inputs with the ``Encoding-Independent Partial Decoding-Fusion Decoding” structure, enabling semantically aligned integration for heterogeneous data. Experimental results on the public dataset demonstrate the superior performance of our model. Future research will focus on enhancing the model's real-time capabilities for seamless integration into Advanced Driver Assistance Systems (ADAS), particularly in edge computing environments.

\ifCLASSOPTIONcaptionsoff
  \newpage
\fi

\bibliographystyle{IEEEtran}

\end{document}